\pdfoutput=1

\documentclass[11pt]{article}

\usepackage[]{ACL2023}

\usepackage{times}
\usepackage{latexsym}

\usepackage[T1]{fontenc}

\usepackage[utf8]{inputenc}

\usepackage{microtype}

\usepackage{inconsolata}

\usepackage{multirow}
\usepackage{booktabs}
\usepackage{color}
\usepackage{makecell}
\usepackage{hyperref}
\usepackage{graphicx}
\usepackage {amsmath}
\usepackage{amsfonts,amssymb} 
\usepackage{algorithm}
\usepackage{algorithmic}
\usepackage{float}

\newcommand*\samethanks[1][\value{footnote}]{\footnotemark[#1]}
\newcommand{\ie}{\textit{i}.\textit{e}.}

\title{UniFine: A Unified and Fine-grained Approach for Zero-shot Vision-Language Understanding}

\author{Rui Sun\textsuperscript{1}\thanks{\indent Equal Contribution}, Zhecan Wang\textsuperscript{1}\samethanks[1], Haoxuan You\textsuperscript{1}\samethanks[1], Noel Codella\textsuperscript{2},
\\
\textbf{Kai-Wei Chang\textsuperscript{3}, Shih-Fu Chang\textsuperscript{1}}
\\
\textsuperscript{1} Columbia University \quad
\textsuperscript{2} Microsoft Research \quad
\textsuperscript{3} University of California, Los Angeles \\
\texttt{\{rs4110, zw2627, hy2612, sc250\}@columbia.edu}\\
\texttt{ncodella@microsoft.com, kwchang@cs.ucla.edu} 
}

\begin{document}
\maketitle
\begin{abstract}

Vision-language tasks, such as VQA, SNLI-VE, and VCR are challenging because they require the model’s reasoning ability to understand the semantics of the visual world and natural language. Supervised methods working for vision-language tasks have been well-studied. However, solving these tasks in a zero-shot setting is less explored. Since Contrastive Language-Image Pre-training (CLIP) has shown remarkable zero-shot performance on image-text matching, previous works utilized its strong zero-shot ability by converting vision-language tasks into an image-text matching problem, and they mainly consider global-level matching (\textit{e.g.}, the whole image or sentence). However, we find visual and textual fine-grained information, \textit{e.g.}, keywords in the sentence and objects in the image, can be fairly informative for semantics understanding. Inspired by this, we propose a unified framework to take advantage of the fine-grained information for zero-shot vision-language learning, covering multiple tasks such as VQA, SNLI-VE, and VCR. Our experiments show that our framework outperforms former zero-shot methods on VQA and achieves substantial improvement on SNLI-VE and VCR. Furthermore, our ablation studies confirm the effectiveness and generalizability of our proposed method. 



\end{abstract}

\section{Introduction}

\begin{figure}[t!]
        \includegraphics[width=0.32\textheight]{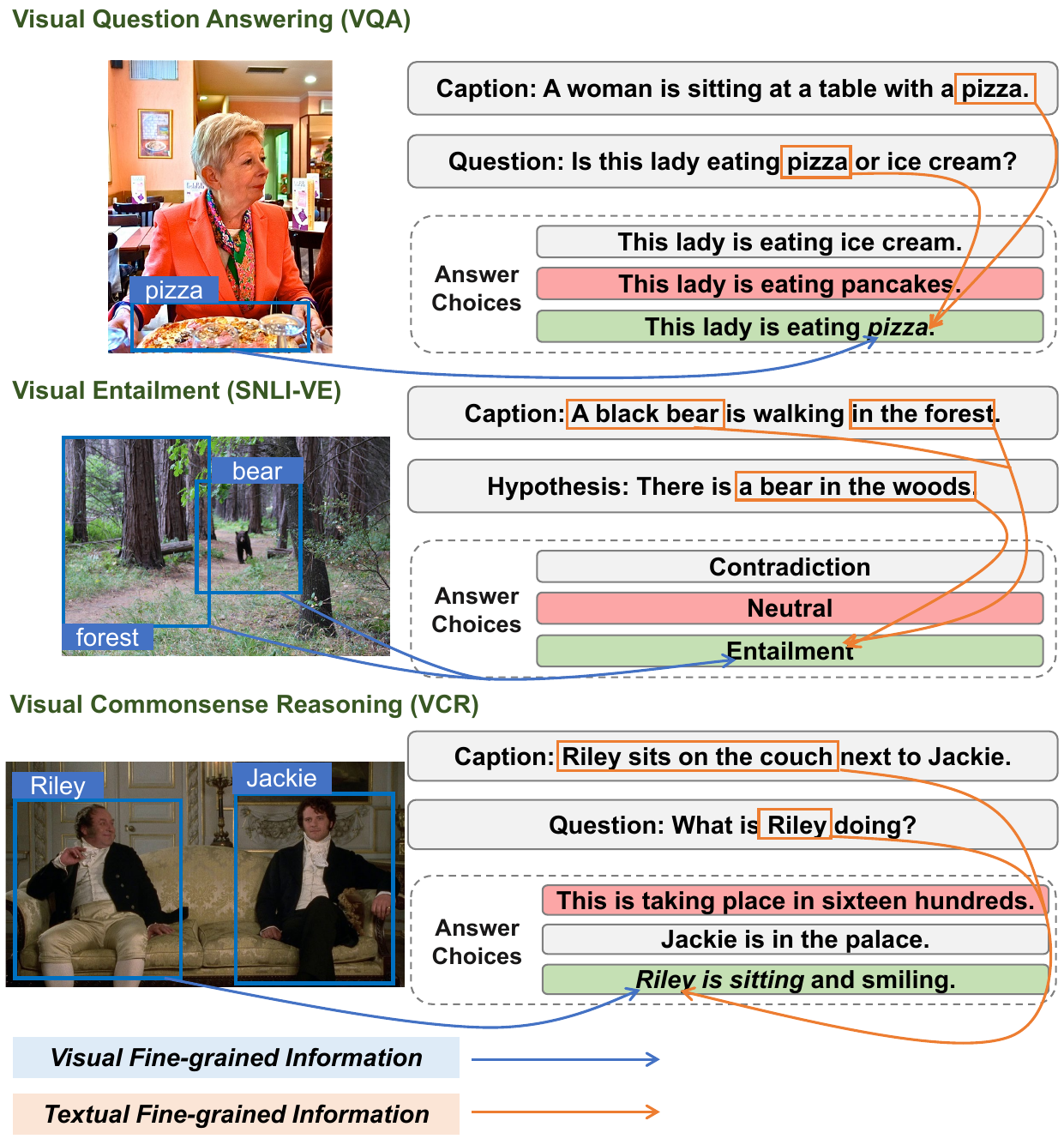}
        \vspace{-5mm}
        \caption{Examples of how fine-grained information is utilized to help CLIP from VQA, SNLI-VE, and VCR. Before extracting the fine-grained information, CLIP gives the wrong answer shown as the \colorbox{red!30}{red} box. With the assistance of visual and textual fine-grained information, CLIP can make the correct decision as the \colorbox{green!20}{green} box shows. (For visualization, only three answer choices are kept in VQA and VCR. And unisex names (Riley and Jackie) are added in VCR.)}
\label{fig:motivation}
\vspace{-6mm}
\end{figure}

VQA \citep{antol2015vqa}, SNLI-VE \citep{xie2019visual}, and VCR \citep{zellers2019recognition} are vision-language tasks, which utilize the text and corresponding image to test a system's cross-modal reasoning ability. These tasks are challenging as requiring models to obtain a joint understanding of visual and textual modality. Nevertheless, they are also meaningful since this capability plays an essential role in daily human-robot interaction, \textit{e.g.}, asking a robot how many people are in the image. Despite the difficulty, a line of work \citep{tan2019lxmert, li2019visualbert, lu2019vilbert, chen2019uniter, su2019vl, li2020oscar} has been dedicated to resolving these vision-language tasks in a supervised setting and obtaining impressive progress. However, these methods all suffer from a significant problem of being costly as they require expert knowledge to collect well-annotated image-text data. On the other hand, zero-shot methods for vision-language tasks can successfully bypass this problem without costly annotations. Unfortunately, limited methods and relatively fewer works have been dedicated to exploring this direction.


Recently, CLIP \citep{radford2021learning} has been proposed to acquire visual concepts using natural language supervision. It jointly trains an image encoder and a text encoder on 400M noisy image-text pairs collected from the Internet by aligning images and texts through a contrastive loss. 

Previous works \citep{song2022clip, subramanian2022reclip, shen2021much, wang2022multimodal} demonstrated that CLIP can achieve strong zero-shot performance of vision-language tasks by converting original tasks into the image-text matching format. However, they mainly consider matching on an instance or global level, \textit{i.e.}, the whole image or sentence, ignoring the significance of fine-grained elements, \textit{e.g.}, keywords in the sentence and objects in the image. Meanwhile, we find these fine-grained elements are important for specific downstream tasks, especially in zero-shot learning. 

For instance, in Fig. \ref{fig:motivation}, CLIP makes three incorrect predictions in three zero-shot vision-language tasks. For VQA, the model infers the wrong object "pancake" for the verb "eating", as it does not capture the details in the image (pizza on the table) and captions (pizza is mentioned). We posit that if we can find a proper solution to navigate the model to focus on these detailed pieces of textual and visual information, the model would likely have a better chance of selecting the correct answer label. This conjecture also seems true and generalizable across multiple zero-shot downstream tasks as shown by the three examples from different vision-language tasks, \textit{i.e.}, VCR, VQA, and SNLI-VE in Fig. \ref{fig:motivation}. Yet, we also recognize potential challenges also exist as those different tasks may differ from many perspectives including the distribution of image categories or scenes, the different semantic focus, and format of text premises between declarative statements and questions, and different task formats in terms of image-text matching or classification.

To overcome these challenges, we first identify two common fundamental steps required to utilize the fine-grained information across different vision-language tasks: 1) Extraction of the fine-grained information from context information, \textit{e.g.,} the extraction of the word "pizza" from the caption in VQA as in Fig. \ref{fig:motivation}. 2) The semantic matching between these extracted fine-grained information and answer choices or hypothesis. Based on these, we propose a unified approach leveraging these two common steps thus it can assist the model to generalize over different vision-language tasks.
For the extractor, we have two branches -- 1) the vision branch and 2) the textual branch. In the vision branch, we employ Faster-RCNN \citep{ren2015faster} to extract object-level information. We select relevant object regions guided by the question in VQA and VCR or hypothesis in SNLI-VE. After that, we concatenate the whole image and its selected image regions and input them into the image encoder of CLIP. For textual information extraction, we exploit rich information from the image caption generated by a recently-developed captioning model OFA \cite{pmlr-v162-wang22al} and question in VQA and VCR or hypothesis in SNLI-VE to boost the zero-shot performance. 

It's noted that although we employ the image caption and question on a sentence level rather than a word level, we compute the cosine similarity between them and answer texts, which means if there are keywords in the answer texts which can be matched in the caption or question, then we will obtain high scores in zero-shot prediction. 

Therefore, it is still a process of fine-grained information extraction. By using fine-grained information, our model outperforms previous methods on zero-shot VQA and we are the first to benchmark zero-shot VCR and SNLI-VE. The experiments confirm the effectiveness of our proposed method.


Our contributions can be summarized as follows:
\begin{itemize}
    \vspace{-2mm}
    \item To the best of our knowledge, we are the first to propose a unified approach based on fine-grained information extraction for zero-shot learning of different vision-language tasks.
    \vspace{-2mm}
    \item Our approach outperforms previous CLIP-based methods for zero-shot learning of VQA and we are the first to study CLIP's zero-shot ability for SNLI-VE and VCR.
    \vspace{-2mm}
    \item The experiments and ablation studies confirm the generalizability of our proposed method and the significance of visual and textual fine-grained information for zero-shot learning of vision-language tasks.
\end{itemize}

\section{Related Work}

\textbf{Vision-language understanding tasks.} Unlike unimodal tasks, vision-language understanding tasks need joint understanding between vision and language, which require a deeper reasoning ability of the system. In VQA \citep{goyal2017making}, given a question, the model needs to understand the details of the corresponding image based on the question to answer correctly. The real images in VQA come from MS COCO\citep{lin2014microsoft} and each of them is paired with a caption in COCO Captions \citep{chen2015microsoft}. For another task VCR \citep{zellers2019recognition}, its semantic focus is different from VQA since it concentrates more on commonsense questions. The model needs to answer the recognitive questions (like VQA) at first, then it is also required to correctly answer the cognitive questions, which are rationales of the choice in the first question. The images in VCR are collected from movie clips. SNLI-VE originated from Stanford Natural Language Inference (SNLI) \citep{bowman2015large}, which is a text entailment (TE) task based on the Flicker30k \citep{young2014image} image captions. It extends TE into the visual domain and it has a different task format from the VQA and VCR because the previous question-answering format is replaced with the hypothesis. Given the image and hypothesis, the model needs to predict whether the image semantically entails the text. The images in SNLI-VE are from Flicker30k with annotated captions. 

\textbf{Vision-language pre-trained models.} Early vision-language pre-trained models \cite{tan2019lxmert, lu2019vilbert, li2019visualbert, chen2019uniter, su2019vl, li2020oscar} utilize cross-modal transformer \citep{vaswani2017attention} pre-trained on well-annotated image-text pairs. Different from these models, contrastive learning frameworks \cite{radford2021learning, pham2021combined, pmlr-v139-jia21b} are trained on noisy image-text pairs crawled from the Internet through contrastive loss, which employs the dot product between visual and textual modality. Due to the large-scale training data, these models acquire rich prior knowledge and show strong zero-shot ability on vision benchmarks like ImageNet \citep{deng2009imagenet}.

\textbf{Vision-language zero-shot learning.} There is a line of work utilizing CLIP to do zero-shot learning for vision-language tasks. ReCLIP \citep{subramanian2022reclip} utilizes CLIP to present a zero-shot method for referring expression comprehension (ReC), which outperforms prior zero-shot ReC approaches. CLIP-ViL \citep{shen2021much} exploits CLIP to do zero-shot VQA by simply concatenating question and answer pair for each question and constructing "question: [question text] answer: [answer text]" as the prompt. Then, they feed the text and image into the text encoder and the image encoder of CLIP, which produces the near-chance level performance. The most relevant work to ours is TAPC \citep{song2022clip}, which manually designs the prompt and leverages T5 \citep{raffel2020exploring}, a large pre-trained text-to-text Transformer, to convert the question-answering problem into the image-text matching task. Then, it employs CLIP's remarkable zero-shot image-text matching ability on VQA, whose results surpass CLIP-ViL by a large margin. However, these works handle different tasks on an instance level rather than fully utilizing the visual and textual fine-grained information (\ie, keywords in the sentence and objects in the image) like ours. Moreover, we can tackle a diverse set of tasks but they just concentrate on one specific task. 



\begin{figure}[t]
        \includegraphics[width=0.31\textheight]{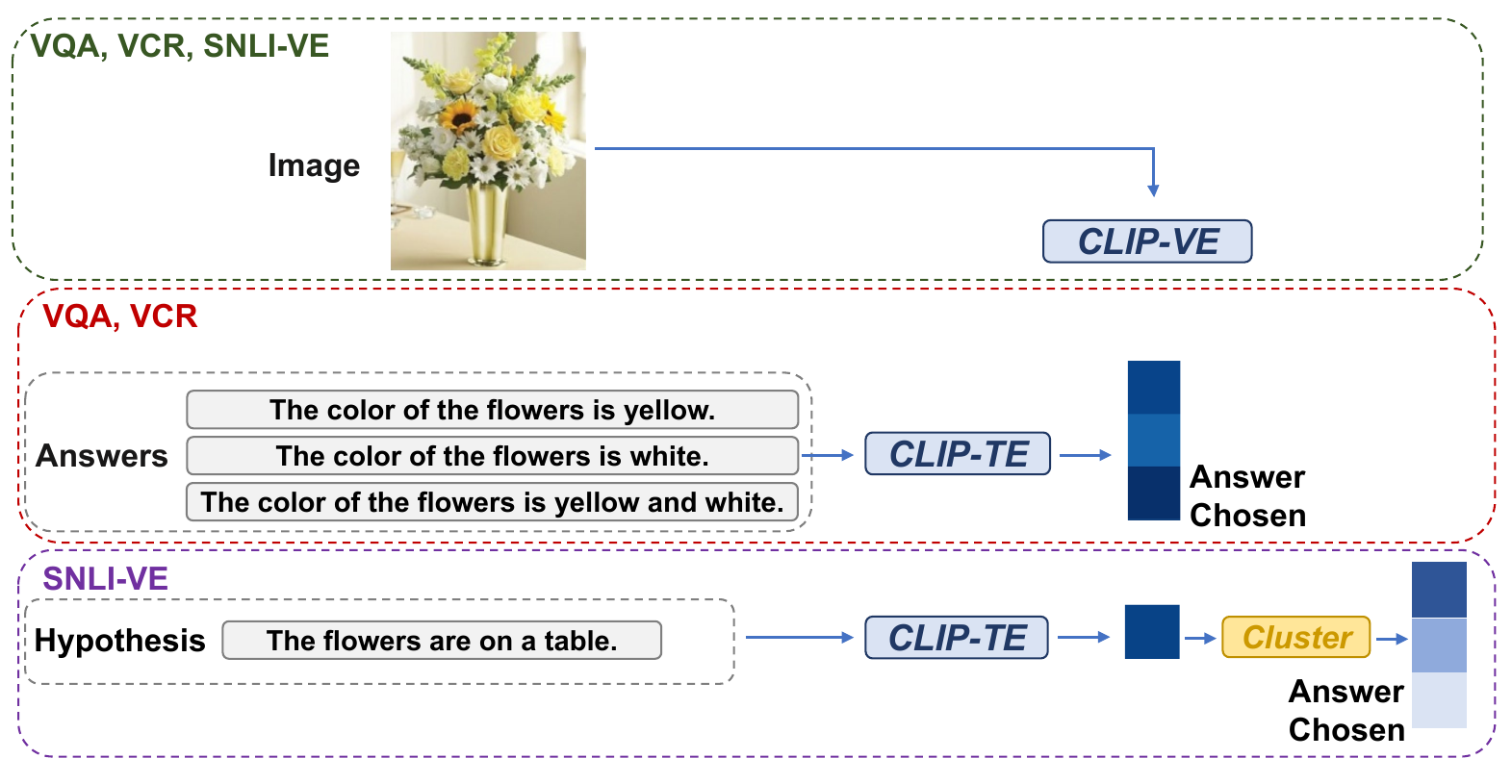}
        \vspace{-6mm}
        \caption{Baseline method of UniFine. (Note: For visualization, only three answer choices are kept in VQA and VCR. CLIP-VE denotes CLIP Visual Encoder, and CLIP-TE denotes CLIP Text Encoder.)}
\label{fig:baseline}
\vspace{-4mm}
\end{figure}

\section{Method}


In this section, we introduce our method for visual and textual fine-grained information extraction to improve zero-shot learning of vision-language tasks including VQA, VCR, and SNLI-VE.

\subsection{Baseline Method}

In the baseline method shown in Fig. \ref{fig:baseline}, we use CLIP to do zero-shot learning of vision-language tasks. CLIP consists of a visual encoder $\mathbb{V}$ (\textit{e.g.}, ResNet \citep{he2016deep} and ViT \citep{dosovitskiy2020image}) and a text encoder $\mathbb{T}$ (\textit{e.g.}, transformer \citep{vaswani2017attention}), where the image and text are processed independently.

Followed by the encoder, there is the dot product (\textit{i.e.}, alignment score) between visual and textual features, \textit{i.e.}, $\mathbb{V}(\mbox{image}) \cdot \mathbb{T}(\mbox{text})$. 
We input the image from VQA, VCR, and SNLI-VE into the CLIP visual encoder. Since there is a difference in task format, answer choices from VQA and VCR and the hypothesis from SNLI-VE are input into the CLIP text encoder. After encoding, we can obtain the alignment score between the image and text. In VQA and VCR, we select the answer with the highest score. In SNLI-VE, there is a clustering process after the dot product, which is demonstrated in Algo. \ref{alg:ve}, and we select the answer with the lowest score.

\begin{figure*}[t]
        \centering
        \includegraphics[width=0.65\textheight]{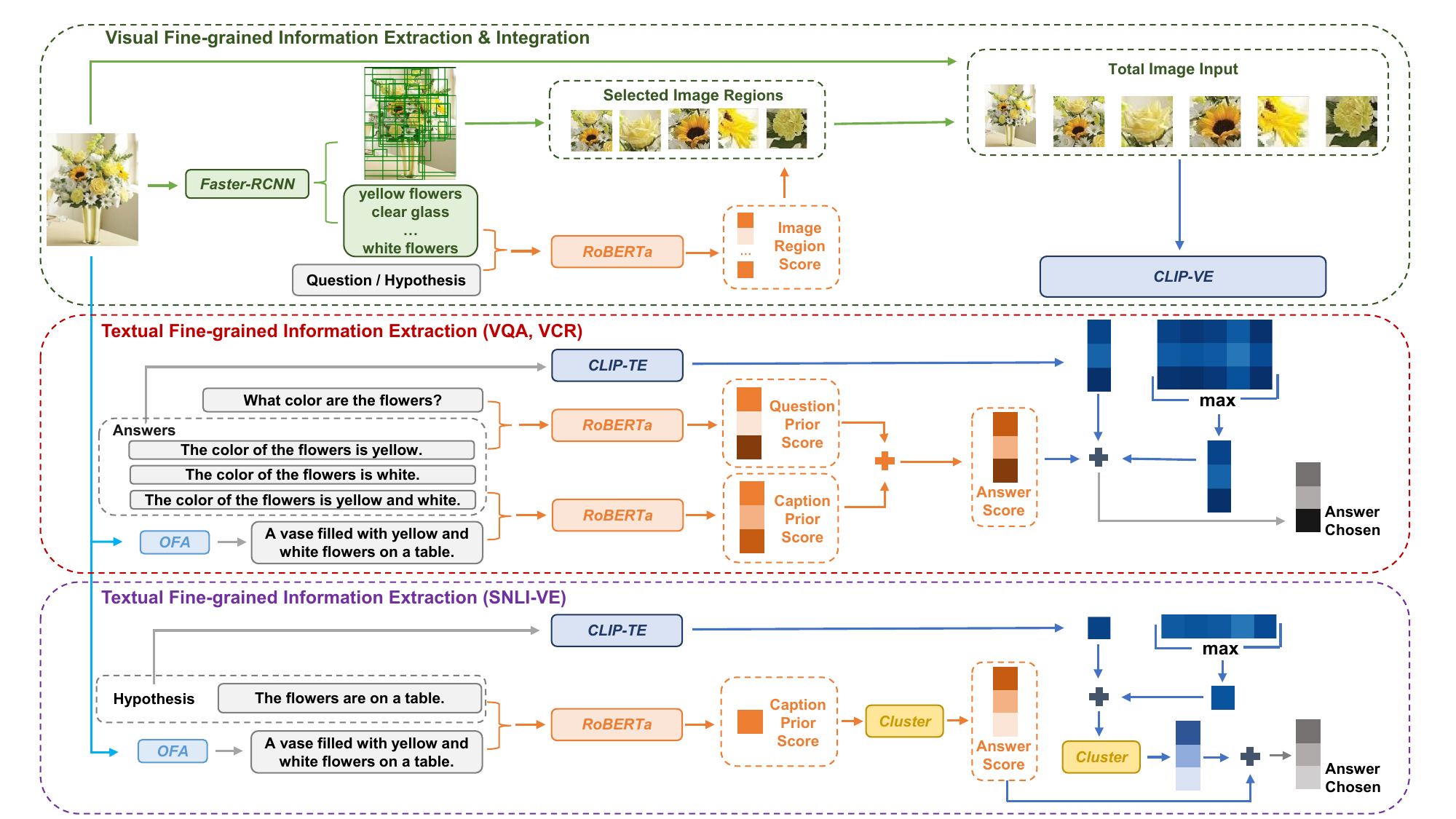}
        \vspace{-5mm}
        \caption{The overview of our proposed UniFine method. Visual fine-grained information is extracted under the guidance of the question (in VQA and VCR) or hypothesis (in SNLI-VE). In addition, textual fine-grained information is extracted by utilizing the question (in VQA and VCR) or hypothesis (in SNLI-VE) and the image caption. (Note: CLIP-VE denotes CLIP Visual Encoder and CLIP-TE denotes CLIP Text Encoder.)}
\label{fig:overview}
\vspace{-5mm}
\end{figure*}

\subsection{Visual fine-grained information extraction}

In visual fine-grained information extraction, we aim to find the related image regions to the question in VQA and VCR or the hypothesis in SNLI-VE since these regions can provide local visual clues to complement the global image. The objects and attributes are detected by Faster-RCNN \citep{ren2015faster}, which is pre-trained on Visual Genome \citep{krishna2017visual} provided by \citet{Anderson2017up-down}. We select the top $N$ relevant image regions ($N$ is a hyperparameter, which will be analyzed in Sec. \ref{sec:visfine}) by image region score (\textit{i.e.}, cosine similarity) between the textual features of the question or hypothesis and the object class\&attribute (\textit{e.g.}, yellow flowers) encoded by RoBERTa \cite{liu2019roberta}:
\begin{equation}
\small
\centering
    \mathop{{\mbox{Top-}N}}_{o_i \in O} \{\mbox{cos}(\mathbb{R}(Query), \mathbb{R}(\{Attr(o_i), Class(o_i)\}))\}
\end{equation}
\noindent where $\mathbb{R}$ is RoBERTa, cos$(,)$ is cosine similarity, $O$ is the set of objects detected by Faster-RCNN, $Attr()$ and $Class()$ are attribute and class of object respectively,  and $Query$ is the question in VQA and VCR or the hypothesis in SNLI-VE. After selection, the global image and selected image regions will be fed into CLIP visual encoder to obtain the encoded feature of each.

\subsection{Textual fine-grained information extraction}

Next, we present how textual fine-grained information is extracted and incorporated into our framework. To be more specific, two types of information are studied: image caption and question. Questions as a prior can narrow down the range of answer candidates and get rid of irrelevant answers.
Image caption can transform the information inside the image into text so that it can be compared with answers in the same domain.   Image captions are generated from the image, but their format is language. 
Thus, we arguably regard image captions as textual fine-grained information. Overcoming the challenge in different formats of vision-language tasks, we introduce a relatively unified way to extract and utilize textual fine-grained information in the zero-shot scenario.

\noindent\textbf{Visual Question Answering: VQA}
\label{sec:vqa}

\noindent Following previous work, we experiment on the validation set of VQAv2 \citep{goyal2017making}. Typically, VQA is regarded as a classification problem and there are 3,129 most frequent answers used for classification. There are 65 types of questions (\textit{e.g.}, \textit{does this} type) and 3 types of answers including \textit{Yes/No}, \textit{Number}, and \textit{Other} in VQAv2. 
 
Although in VQA, each image is paired with a ground truth caption from MS COCO, we still choose to use OFA, a SOTA model of image captioning, to generate the caption given the image, because not every dataset is annotated with ground truth captions and we would like to make our method generalizable.

As \citet{shen2021much} shows, directly inputting the concatenation of the question and answer into CLIP will lead to near-chance level performance. In addition, there are more than 3,000 answer candidates in VQAv2, which will largely slow down the inference speed of zero-shot VQA with all answers input into CLIP. To bypass that, we utilize an answer-filtering method to downsize the number of answer choices inspired by \citet{song2022clip}. 


Following \citet{song2022clip}, we first convert the question-answering format into declarative templates with the \textit{<extra\_id\_0>} token by T5 low-shot demonstration. Then, templates with \textit{<extra\_id\_0>} token are input into the T5 and we obtain the plausibility of each answer candidate based on T5 output probability. Next, we select the top $K$ answers. More details can be found at Sec. \ref{sec:answerfiltering}. 

In this way, we can downsize the number of answers in VQA. There are three different answer types in VQA, which are processed differently in the answer filtering process. For \textit{Yes/No} type, we treat it as a binary classification problem. For \textit{Number} type, since its answers are highly related to numerical answers in the 3,129 most frequent answers set, we heuristically filter 285 numerical answers from 3,129 answers before answer filtering. As for \textit{Other} type, we preserve the original answer candidates without filtering.

After obtaining top $K$ filtered answers, on one hand, they will be sent to the CLIP text encoder and dot-product with image features will be calculated, denoted as CLIP alignment score $S_{\mbox{\tiny CLIP}}$. On the other hand, we will calculate the question prior score $S_{\mbox{\tiny Question}}$ (\textit{i.e.}, cosine similarity between textual features, encoded by RoBERTa, of the question and answers) and the caption prior score $S_{\mbox{\tiny Caption}}$ (\textit{i.e.}, cosine similarity between textual features, encoded by RoBERTa, of image caption generated by OFA and answers). The whole process can be summarized as the following equations:
\begin{equation}
\begin{aligned}
    & S_{\mbox{\tiny CLIP}} = \mathbb{T}(A) \cdot \mathbb{V}(I) \\
    & S_{\mbox{\tiny Question}} = \mbox{cos}(\mathbb{R}(Q), \mathbb{R}(A)) \\
    & S_{\mbox{\tiny Caption}} = \mbox{cos}(\mathbb{R}(\mathbb{O}(I), \mathbb{R}(A))
\end{aligned}
\end{equation}
where $\mathbb{V}$ and $\mathbb{T}$ are image and text encoders of CLIP, $\mathbb{R}$ is RoBERTa, $\mathbb{O}$ is OFA, and cos$(,)$ means cosine similarity. $I$ denotes images including one global image $I_g$ and $N$ selected image regions $\{I_l \in Reg\}$. $Q$ and $A$ correspond to the question and its top $K$ filtered answers. $\mathbb{O}(I)$ means image caption generated by OFA.

In the end, all scores are ensembled. We select the answer with the highest score as zero-shot prediction result:
\begin{equation}\normalsize
\small
\centering
\begin{aligned}
    \arg \max_i & \{ S_{\mbox{\tiny CLIP}}(A_i, I_g) + k_1 \cdot \max_{I_l \in Reg}\{S_{\mbox{\tiny CLIP}}(A_i, I_l)\} \\
    & + k_2 \cdot S_{\mbox{\tiny Question}}(Q, A_i) + k_3 \cdot S_{\mbox{\tiny Caption}}(I_g, A_i)\},
\end{aligned}
\end{equation}
\noindent where   $k_1$, $k_2$, and $k_3$ are hyperparameters.

\noindent\textbf{Visual Commonsense Reasoning: VCR}
\label{sec:vcr}


\noindent VCR is similar to VQA since both of them are in question-answering formats. However, there are only four answer choices per question, which means we don't need to do answer filtering. \textit{Q2A} and \textit{QA2R} are two subtasks of VCR. \textit{Q2A} is similar to VQA in that there is only one question per sample. So the process of \textit{Q2A} is the same as VQA except for omitting answer filtering. \textit{QA2R} aims to dig out the rationale why one correct answer is chosen in \textit{Q2A} question. Since there is no question text in \textit{QA2R} and the correct answer is provided, we directly utilize the correct answer as the question text. Other procedures in \textit{QA2R} are the same as \textit{Q2A}.

\noindent\textbf{Visual Entailment: SNLI-VE}
\label{sec:ve}

\noindent The task format of SNLI-VE is different from VQA and VCR. For each sample, only one image premise $I$ and one hypothesis $H$ are given, without answer candidates. It is a three-way classification problem, aiming to predict the relation between the image premise and hypothesis text into one of three classes: \textit{Entailment}, \textit{Contradiction}, and \textit{Neutral}. 

Since there are no answer candidates, we cannot directly compare CLIP alignment scores of answers to select the best answer, as in VQA and VCR.  To tackle that, we compute the CLIP alignment scores between image and hypothesis of each sample in whole evaluation set, and cluster those scores into three clusters with three centroids. We rank the centroids from high to low and sequentially treat them as entailment centroid $C_{\mbox{\tiny CLIP}}^{e}$, neutral centroid $C_{\mbox{\tiny CLIP}}^{n}$ and contradictory centroid $C_{\mbox{\tiny CLIP}}^{c}$. The detail of clustering can be found in Algo. \ref{alg:ve}. It's noted that, to make cluster centroids meaningful, an assumption is required: three relationships are uniformly distributed in the evaluation dataset. That assumption is true in SNLI-VE but not guaranteed in other less-calibrated datasets. We can measure how close $S_{\mbox{\tiny CLIP}}$ of each sample is to each centroid:
\begin{equation}
\small
\begin{aligned}
     &  Dis (C_{\mbox{\tiny CLIP}}^{i}, S_{\mbox{\tiny CLIP}})= \\
     & \|C_{\mbox{\tiny CLIP}}^{i} - 
      ( S_{\mbox{\tiny CLIP}}(H, I_g) + k_1 \cdot \max_{I_l \in Reg}\{S_{\mbox{\tiny CLIP}}(H, I_l)\}) \| \\
\end{aligned}
\end{equation}
where centroid $C_{\mbox{\tiny CLIP}}^{i} \in \{C_{\mbox{\tiny CLIP}}^{e}, C_{\mbox{\tiny CLIP}}^{n}, C_{\mbox{\tiny CLIP}}^{c}\}$. 

Besides the CLIP alignment score comparison, we can obtain the caption prior score $S_{\mbox{\tiny Caption}}(I, H)$ using the image caption generated by OFA. Same as above, we also use the clustering method in Algo. \ref{alg:ve}, with only changing CLIP score to caption score, to get three centroids $\{C_{\mbox{\tiny Caption}}^{e}, C_{\mbox{\tiny Caption}}^{n}, C_{\mbox{\tiny Caption}}^{c}\}$. And we measure how close $S_{\mbox{\tiny Caption}}$ of each sample is to each centroid:
\begin{equation}
\small
\begin{aligned}
     &  Dis (C_{\mbox{\tiny Caption}}^{i}, S_{\mbox{\tiny Caption}})= \|C_{\mbox{\tiny Caption}}^{i} -  S_{\mbox{\tiny Caption}}(I, H) \| \\
\end{aligned}
\end{equation}
 It's noted that due to the lack of answer candidates, we can't get the question prior score $S_{\mbox{\tiny Question}}$. In the end, we ensemble two distances and predict the relationship by picking the closest centroid:
 \begin{equation}
 \small
\arg \min_i \{Dis (C_{\mbox{\tiny CLIP}}^{i}, S_{\mbox{\tiny CLIP}}) + k_2 \cdot  Dis (C_{\mbox{\tiny Caption}}^{i}, S_{\mbox{\tiny Caption}}) \}
 \end{equation}

\vspace{-3mm}

\section{Experiments}


In this section, we will talk about benchmark comparison first to show our strong performance. Then, we conduct extensive ablation studies to confirm the effectiveness of fine-grained information. 

\subsection{Experimental setup}

\textbf{Datasets.} We analyze three vision-language tasks in our paper. For each of them, we utilize the validation set of VQAv2 \citep{goyal2017making}, VCR \citep{zellers2019recognition}, and SNLI-VE \citep{xie2019visual}. More details about the validation set can be found in Sec. \ref{sec:statistics}. In VQAv2, we employ \textit{vqa scores} to evaluate the model. In VCR and SNLI-VE, we use the accuracy of the validation set for evaluation.

\noindent \textbf{Models.} The core component of our method is CLIP \footnote{\href{https://github.com/openai/CLIP}{https://github.com/openai/CLIP}}. There are different variants of CLIP since we can use different models to act as the image or text encoder. Following previous work, we leverage CLIP Res50x16 and CLIP ViT-B/16 in VQA for comparison. Since we are the first to evaluate CLIP's zero-shot ability in SNLI-VE and VCR, there is no need for us to compare them with prior work. So we just exploit CLIP ViT-B/16 in VCR and SNLI-VE. We believe the scale of the model will have a big impact on the result, so we also utilize CLIP ViT-L/14@336px in VQA, VCR, and SNLI-VE to see how much improvement can be obtained by using a larger model. In addition to CLIP, we also use T5-large \footnote{\href{https://huggingface.co/models}{https://huggingface.co/models}} for task format conversion, OFA-base \footnote{\href{https://github.com/OFA-Sys/OFA}{https://github.com/OFA-Sys/OFA}} for image captioning, RoBERTa-large \footnote{\href{https://github.com/UKPLab/sentence-transformers}{https://github.com/UKPLab/sentence-transformers}} for the following calculation of cosine similarity, and Faster-RCNN \footnote{\href{https://github.com/peteanderson80/bottom-up-attention}{https://github.com/peteanderson80/bottom-up-attention}} for object detection.

\subsection{Benchmark comparison}


\begin{table}[t]\scriptsize
\setlength\tabcolsep{4pt} 
\renewcommand{\arraystretch}{1.2} 
\centering
\begin{tabular}{l|cccc} 
    \toprule
    \multirow{2}{*}{Methods} & \multicolumn{4}{c}{VQA Answer Types} \\
    & Yes/No & Number & Other & All \\
    \midrule
    \textbf{CLIP-ViL} & & & &  \\
    w/ CLIP$_{\mbox{\tiny Res50x16}}$ & 56.16 & 9.76 & 1.39 & 23.07 \\
    w/ CLIP$_{\mbox{\tiny ViT-B/16}}$ & 53.89 & 7.67 & 0.70 & 21.40 \\
    \midrule
    \textbf{TAP-C* (Baseline)} & & & & \\ 
    w/ CLIP$_{\mbox{\tiny Res50x16}}$ & 68.9(\textbf{71.7}) & 25.9(18.7) & 16.7(18.2) & 37.5(38.4) \\
    w/ CLIP$_{\mbox{\tiny ViT-B/16}}$ & 68.6(71.4) & 25.4(21.0) & 16.7(18.6) & 37.3(38.7) \\
    \midrule
    \textbf{UniFine-Base (Ours)} & & & & \\
    w/ CLIP$_{\mbox{\tiny Res50x16}}$ & 69.69 & 29.61 & 19.85 & 39.87 \\
    w/ CLIP$_{\mbox{\tiny ViT-B/16}}$ & 69.49 & 29.34 & 20.15 & 39.91 \\
    \midrule
    \textbf{UniFine-Large (Ours)} & & & & \\
    w/ CLIP$_{\mbox{\tiny ViT-L/14@336px}}$ & 70.36 & \textbf{29.95} & \textbf{20.19} & \textbf{40.33} \\
    \midrule
    \textbf{Random} & 50.00 & - & - & 18.80 \\
    \bottomrule
\end{tabular}
\vspace{-1mm}
\caption{\label{tab:vqa} 
Zero-shot VQAv2 results on the validation set. * denotes our reimplementation. Reported results from TAP-C are in the bracket.
}
\vspace{-2mm}
\end{table}

\begin{table}[t]\scriptsize
\setlength\tabcolsep{4pt} 
\renewcommand{\arraystretch}{1.2} 
\centering
\begin{tabular}{l|cccc|cc} 
    \toprule
    \multirow{2}{*}{Methods} & \multicolumn{4}{c|}{SNLI-VE Answer Types} & \multicolumn{2}{c}{VCR Tasks} \\
    & C & N & E & All & Q2A & QA2R \\
    \midrule
    \textbf{Baseline} & & & &  \\
    w/ CLIP$_{\mbox{\tiny ViT-B/16}}$ & 67.59 & 18.66 & \textbf{55.92} & 47.37 & 53.24 & 46.51 \\
    \midrule
    \textbf{UniFine-Base} & & & & \\
    w/ CLIP$_{\mbox{\tiny ViT-B/16}}$ & 68.08 & 28.55 & 51.67 & 49.41 & 54.97 & 50.72 \\
    \midrule
    \textbf{UniFine-Large} & & & & \\
    w/ CLIP$_{\mbox{\tiny ViT-L/14@336px}}$ & \textbf{68.29} & \textbf{29.57} & 52.68 & \textbf{50.16} & \textbf{58.48} & \textbf{51.88} \\
    \midrule
    \textbf{Supervised} & & & & \\
    w/ EVE-Image & 71.04 & 70.55 & 73.10 & 71.56 & \multicolumn{2}{c}{-} \\
    w/ R2C & \multicolumn{4}{c|}{-} & 63.8 & 67.2 \\
    \midrule
    \textbf{Random} & - & - & - & 33.33 & 25.00 & 25.00 \\
    \bottomrule
\end{tabular}
\vspace{-1mm}
\caption{\label{tab:vevcr} 
Zero-shot SNLI-VE and VCR results on the validation set. (C: Contradiction, N: Neutral, E: Entailment)
}
\vspace{-5mm}
\end{table}

\textbf{VQA.} Results of zero-shot VQA are reported in Tab. \ref{tab:vqa}. For a fair comparison, we compare our method with two CLIP-based methods. We choose TAP-C \citep{song2022clip} as our baseline. Since the author didn't release the code, we reimplement it from scratch. After reimplementation, we obtain a lower score than TAP-C. There might be some reasons like differences in specific prompt design and answer filtering process making our result different from the original one. Although our reimplemented results are lower than the reported ones, we can surpass TAP-C after extracting and exploiting visual and textual fine-grained information. Compared to our reimplemented results, our method can elevate the performance of all answer types. By using a larger CLIP model, we can achieve better performance. Our best performance can surpass the reimplemented and TAP-C result by 2.83\% and 1.63\%. Currently, our method outperforms previous CLIP-based methods for zero-shot VQA.

\begin{table*}[t]\scriptsize
\setlength\tabcolsep{1pt} 
\renewcommand{\arraystretch}{1.4} 
\centering
\begin{tabular}{l|cc|cccccccc} 
    \toprule
    Methods & \multicolumn{2}{c|}{VLU Tasks} & Baseline & w/ Question & \makecell[c]{w/ Image Region \\ (Faster-RCNN)} & \makecell[c]{w/ Image Caption \\ (OFA)} & \makecell[c]{w/ All \\ (Generation)} & \makecell[c]{w/ Image Caption \\ (GT)} & \makecell[c]{w/ Image Region \\ (GT)} & \makecell[c]{w/ All \\ (GT)} \\
    \midrule
    \multirow{10}{*}{\makecell[l]{UniFine-Base \\ w/ CLIP$_{\mbox{\tiny Base}}$}} & \multirow{4}{*}{VQA} & Yes/No & 68.61 & 68.67 \textcolor{cyan}{[+0.06]} & 68.80 \textcolor{cyan}{[+0.19]} & 69.40 \textcolor{cyan}{[+0.79]} & 69.49 \textcolor{cyan}{[+0.88]} & 69.32 \textcolor{cyan}{[+0.71]} & - & 69.54 \textcolor{cyan}{[+0.93]} \\
    & & Number & 25.36 & 26.22 \textcolor{cyan}{[+0.86]} & 26.21 \textcolor{cyan}{[+0.85]} & 28.99 \textcolor{cyan}{[+3.63]} & 29.34 \textcolor{cyan}{[+3.98]} & 28.52 \textcolor{cyan}{[+3.16]} & - & 28.88 \textcolor{cyan}{[+3.52]} \\
    & & Other & 16.65 & 17.61 \textcolor{cyan}{[+0.96]} & 19.00 \textcolor{cyan}{[+2.35]} & 18.26 \textcolor{cyan}{[+1.61]} & 20.15 \textcolor{cyan}{[+3.50]} & 18.03 \textcolor{cyan}{[+1.38]} & - & 20.04 \textcolor{cyan}{[+3.39]} \\
    & & All & 37.33 & 37.94 \textcolor{cyan}{[+0.61]} & 38.67 \textcolor{cyan}{[+1.34]} & 38.90 \textcolor{cyan}{[+1.57]} & 39.91 \textcolor{cyan}{[+2.58]} & 38.69 \textcolor{cyan}{[+1.36]} & - & 39.81 \textcolor{cyan}{[+2.48]} \\
    \cmidrule{2-11}
    & \multirow{4}{*}{SNLI-VE} & C & 67.59 & - & 68.59 \textcolor{cyan}{[+1.00]} & 66.87 \textcolor{magenta}{[-0.72]} & 68.08 \textcolor{cyan}{[+0.49]} & 69.41 \textcolor{cyan}{[+1.82]} & - & 69.67 \textcolor{cyan}{[+2.08]} \\
    & & N & 18.66 & - & 20.94 \textcolor{cyan}{[+2.28]} & 19.84 \textcolor{cyan}{[+1.18]} & 28.55 \textcolor{cyan}{[+9.89]} & 24.87 \textcolor{cyan}{[+6.21]} & - & 25.50 \textcolor{cyan}{[+6.84]} \\
    & & E & 55.92 & - & 53.13 \textcolor{magenta}{[-2.79]} & 56.43 \textcolor{cyan}{[+0.51]} & 51.67 \textcolor{magenta}{[-4.25]} & 63.03 \textcolor{cyan}{[+7.11]} & - & 62.90 \textcolor{cyan}{[+6.98]} \\
    & & All & 47.37 & - & 47.54 \textcolor{cyan}{[+0.17]} & 47.68 \textcolor{cyan}{[+0.31]} & 49.41 \textcolor{cyan}{[+2.04]} & 52.42 \textcolor{cyan}{[+5.05]} & - & 52.66 \textcolor{cyan}{[+5.29]}\\
    \cmidrule{2-11}
    & \multirow{2}{*}{VCR} & Q2A & 53.24 & 54.60 \textcolor{cyan}{[+1.36]} & 53.93 \textcolor{cyan}{[+0.69]} & 53.35 \textcolor{cyan}{[+0.11]} & 54.97 \textcolor{cyan}{[+1.73]} & 53.42 \textcolor{cyan}{[+0.18]} & 53.79 \textcolor{cyan}{[+0.55]} & 54.72 \textcolor{cyan}{[+1.48]} \\
    & & QA2R & 46.51 & 50.10 \textcolor{cyan}{[+3.59]} & 47.66 \textcolor{cyan}{[+1.15]} & 46.79 \textcolor{cyan}{[+0.28]} & 50.72 \textcolor{cyan}{[+4.21]} & 46.60 \textcolor{cyan}{[+0.09]} & 47.02 \textcolor{cyan}{[+0.51]} & 50.16 \textcolor{cyan}{[+3.65]}\\
    \midrule
    \multirow{10}{*}{\makecell[l]{UniFine-Large \\ w/ CLIP$_{\mbox{\tiny Large}}$}} & \multirow{4}{*}{VQA} & Yes/No & 69.38 & 69.42 \textcolor{cyan}{[+0.04]} & 69.75 \textcolor{cyan}{[+0.37]} & 70.04 \textcolor{cyan}{[+0.66]} & 70.36 \textcolor{cyan}{[+0.98]} & 70.16 \textcolor{cyan}{[+0.78]} & - & 70.41 \textcolor{cyan}{[+1.03]} \\
    & & Number & 28.44 & 28.67 \textcolor{cyan}{[+0.23]} & 28.64 \textcolor{cyan}{[+0.20]} & 29.48 \textcolor{cyan}{[+1.04]} & 29.95 \textcolor{cyan}{[+1.51]} & 29.32 \textcolor{cyan}{[+0.88]} & - & 29.72 \textcolor{cyan}{[+1.28]} \\
    & & Other & 16.74 & 17.35 \textcolor{cyan}{[+0.61]} & 19.03 \textcolor{cyan}{[+2.29]} & 17.95 \textcolor{cyan}{[+1.21]} & 20.19 \textcolor{cyan}{[+3.45]} & 17.74 \textcolor{cyan}{[+1.00]} & - & 20.06 \textcolor{cyan}{[+3.32]} \\
    & & All & 38.07 & 38.41 \textcolor{cyan}{[+0.34]} & 39.36 \textcolor{cyan}{[+1.29]} & 39.05 \textcolor{cyan}{[+0.98]} & 40.33 \textcolor{cyan}{[+2.26]} & 38.97 \textcolor{cyan}{[+0.90]} & - & 40.26 \textcolor{cyan}{[+2.19]} \\
    \cmidrule{2-11}
    & \multirow{4}{*}{SNLI-VE} & C & 67.57 & - & 68.44 \textcolor{cyan}{[+0.87]} & 66.69 \textcolor{magenta}{[-0.88]} & 68.29 \textcolor{cyan}{[+0.72]} & 69.37 \textcolor{cyan}{[+1.80]} & - & 70.30 \textcolor{cyan}{[+2.73]} \\
    & & N & 25.17 & - & 20.97 \textcolor{magenta}{[-4.20]} & 29.05 \textcolor{cyan}{[+3.88]} & 29.57 \textcolor{cyan}{[+4.40]} & 25.71 \textcolor{cyan}{[+0.54]} & - & 25.44 \textcolor{cyan}{[+0.27]} \\
    & & E & 51.57 & - & 55.14 \textcolor{cyan}{[+3.57]} & 52.70 \textcolor{cyan}{[+1.13]} & 52.68 \textcolor{cyan}{[+1.11]} & 62.84 \textcolor{cyan}{[+11.27]} & - & 62.84 \textcolor{cyan}{[+11.27]} \\
    & & All & 48.05 & - & 48.15 \textcolor{cyan}{[+0.10]} & 49.46 \textcolor{cyan}{[+1.41]} & 50.16 \textcolor{cyan}{[+2.11]} & 52.62 \textcolor{cyan}{[+4.57]} & - & 52.84 \textcolor{cyan}{[+4.79]} \\
    \cmidrule{2-11}
    & \multirow{2}{*}{VCR} & Q2A & 56.92 & 58.12 \textcolor{cyan}{[+1.20]} & 57.26 \textcolor{cyan}{[+0.34]} & 57.11 \textcolor{cyan}{[+0.19]} & 58.48 \textcolor{cyan}{[+1.56]} & 57.01 \textcolor{cyan}{[+0.09]} & 57.37 \textcolor{cyan}{[+0.45]} & 58.29 \textcolor{cyan}{[+1.37]} \\
    & & QA2R & 48.06 & 51.20 \textcolor{cyan}{[+3.14]} & 49.31 \textcolor{cyan}{[+1.25]} & 48.44 \textcolor{cyan}{[+0.38]} & 51.88 \textcolor{cyan}{[+3.82]} & 48.16 \textcolor{cyan}{[+0.10]} & 48.44 \textcolor{cyan}{[+0.38]} & 51.30 \textcolor{cyan}{[+3.24]} \\
    \bottomrule
\end{tabular}
\vspace{-1mm}
\caption{\label{tab:ablation} 
Ablation studies of zero-shot VQA, SNLI-VE, and VCR. (C: Contradiction, N: Neutral, E: Entailment, GT: Ground Truth, CLIP$_{\mbox{\tiny Base}}$: CLIP$_{\mbox{\tiny ViT-B/16}}$, CLIP$_{\mbox{\tiny Large}}$: CLIP$_{\mbox{\tiny ViT-L/14@336px}}$)
}
\vspace{-4mm}
\end{table*}

\textbf{SNLI-VE.} We report the results of SNLI-VE in Tab. \ref{tab:vevcr}. By using the baseline method, we can get an accuracy of 47.37\% in all categories, which is 14.04\% higher than random performance. This result reveals that our baseline method is strong and it confirms CLIP's zero-shot ability in SNLI-VE. By extracting fine-grained information and upscaling the model, we can increase accuracy by 2.79\% at most. For each answer type, \textit{Neutral} type increases the most (+10.91\%) and \textit{Entailment} type decreases by 3.24\%. We need to note that \textit{Neutral} type is more complex than \textit{Entailment} and \textit{Contradiction} since this type is not as clear as the other two types requiring a model's deeper reasoning ability. The improvement in \textit{Neutral} type shows the significance of fine-grained information. As for the decrement in \textit{Entailment} type, it is likely due to the deficiency of our clustering method, which should be improved in the future. Since there is no CLIP-based zero-shot method for SNLI-VE before, we choose the supervised method EVE-Image from SNLI-VE paper \citep{xie2019visual} for comparison. Although the overall performance is still not comparable to the supervised method, our result of \textit{Contradiction} type is approaching EVE-Image.

\textbf{VCR.} The results of VCR are reported in Tab. \ref{tab:vevcr}. We carry out experiments in two VCR subtasks, namely \textit{Q2A} and \textit{QA2R}. Compared to the random performance of \textit{Q2A} and \textit{QA2R}, our baseline method can improve by 28.24\% and 21.51\% respectively. The improvement confirms CLIP's strong zero-shot ability for VCR. By extracting fine-grained information and using a larger model, we can improve the baseline by 5.24\% and 5.37\% at most, which proves the effectiveness of our proposed method. There is no prior CLIP-based method for zero-shot VCR so we select the supervised model R2C, proposed in VCR paper \citep{zellers2019recognition}, for comparison. Although we cannot surpass the supervised model, the result of \textit{Q2A} is approaching R2C and our results are competitive.

\subsection{Ablation studies} \label{sec:ablation}

In this section, we will analyze every important component of our proposed method. In Tab. \ref{tab:ablation}, we can see all of the fine-grained (FG) information can help zero-shot learning and all fine-grained (FG) information combined together can bring more improvement.

\noindent\textbf{Textual FG Information - Question: }By adding the question prior information, we can see it can help VCR the most. We think the first reason is the question and answer in VCR are longer and more complex than the other two datasets. Consequently, the question and answers can provide more useful and richer information in zero-shot inference. Secondly, the correct answer is likely to have more overlap with the question. Plus, we can observe that question doesn't help a lot in VQA \textit{Yes/No} answer type since this is a binary classification problem and a large number of questions are like \textit{"Is this A or B?"} type, which cannot provide more useful information to zero-shot prediction.

\begin{figure}[t!]
        \centering
        \includegraphics[width=0.27\textheight]{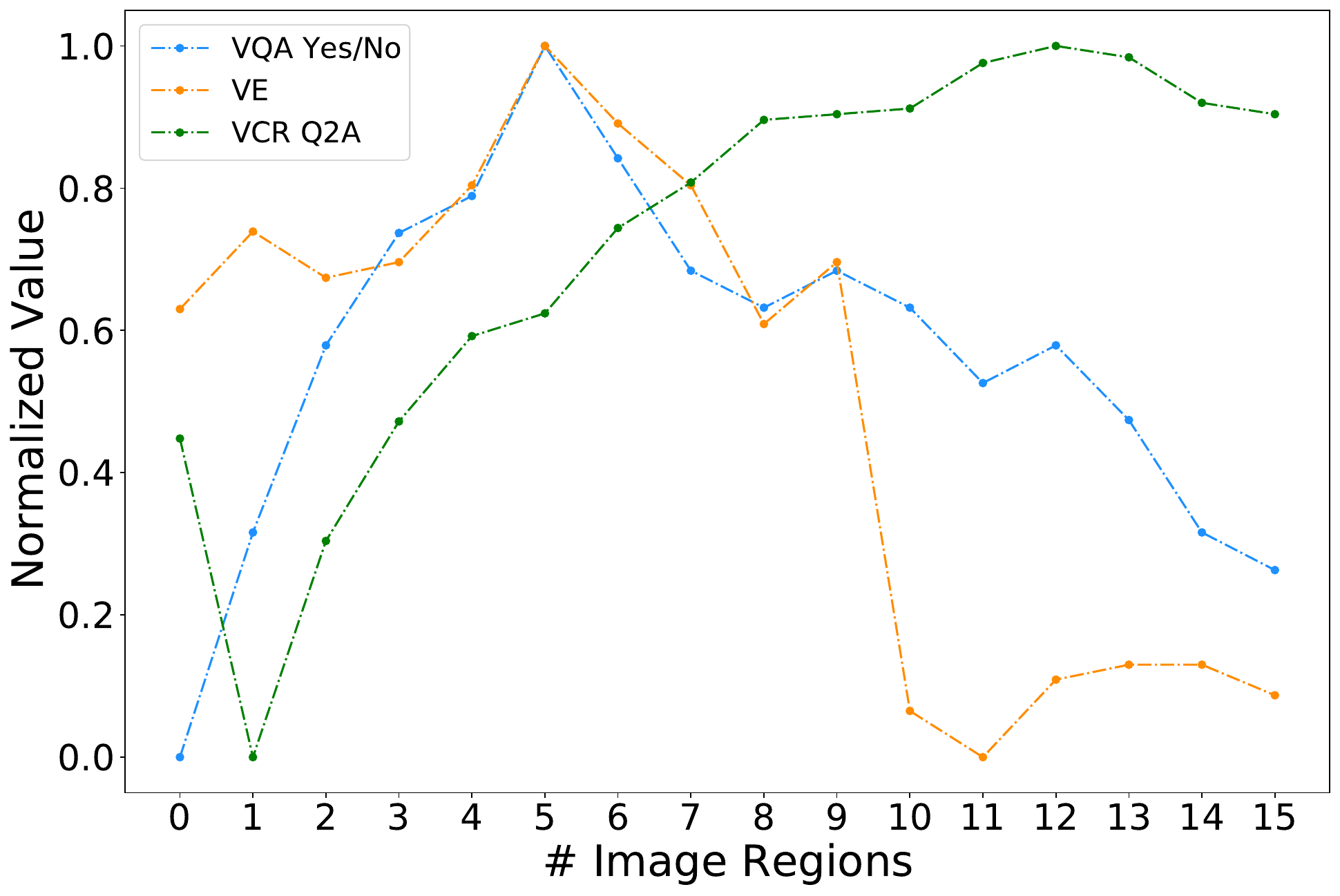}
        \vspace{-1mm}
        \caption{Normalized value to the number of image regions.}
\label{fig:imageregions}
\vspace{-4mm}
\end{figure}

\noindent\textbf{Visual FG Information - Image Region: } We can observe that the image region can largely improve the performance of \textit{Other} answer type in VQA because the questions of this type tend to query the details of the image. And image regions can provide finer details to zero-shot inference. At the same time, we also find that the image region cannot help SNLI-VE much. We think SNLI-VE concentrates more on the global image thus image regions can't help a lot.

\noindent\textbf{Textual FG Information - Image Caption: }In Tab. \ref{tab:ablation}, we can observe that the image caption can better assist the \textit{Number} and \textit{Other} answer type in VQA. For \textit{Number} type, we think the image caption may contain numerical information which aids zero-shot prediction of \textit{Number} type. Since there are a large number of questions in \textit{Other} type, they will cover diverse question types, some of which may focus on information on an instance level. Normally, the image caption captures the instance-level information, so it can help VQA \textit{Other} answer type. We can also notice that using image captions may hurt some categories of SNLI-VE, we think this result may suffer from the quality of the generated caption.

\noindent\textbf{Generation vs. Ground Truth: }Since not every dataset is well human-annotated, we employ these two settings to test the generalizability of our proposed method. In the generation setting, we generate image captions by OFA and detect objects by Faster-RCNN. In the ground truth setting, as mentioned above, there are ground truth captions paired with images in VQA and SNLI-VE. For VCR, images are not paired with human-annotated caption texts. However, 68\% images of VCR validation set are the same as images in VisualCOMET \citep{park2020visualcomet} and VisualCOMET is paired with the ground truth caption. Thus, we directly leverage captions from VisualCOMET in VCR. Although images in VCR are not paired with captions, they are annotated with ground truth bounding boxes, so we have a ground truth image region experiment for VCR. However, VQA and SNLI-VE are not annotated with ground truth bounding boxes. As Tab. \ref{tab:ablation} shows, we can conclude that our method can work well in a situation without many annotations because we achieve similar performance in generation and ground truth scenarios, which confirms the generalizability of our proposed method.

\noindent\textbf{Model Scale: }We believe that the model scale will affect the final result since larger models are able to better process visual and textual information. In our experiments, we mainly focus on two variants of CLIP, namely  CLIP ViT-B/16 and CLIP ViT-L/14@336px. We also carry out experiments on CLIP Res50x16 in VQA task, which can be found in Tab. \ref{tab:ablation2}. We can observe that larger models can elevate the performance and all of our best results are achieved by using CLIP ViT-L/14@336px.

\label{sec:visfine}
\noindent\textbf{Number of Image Regions: }In this subsection, we would like to see how selected $N$ image regions affect the zero-shot performance of different vision-language tasks. For convenience, we select \textit{Yes/No} answer type of VQA, SNLI-VE, and \textit{Q2A} task of VCR to carry out experiments. Full results are reported in Tab. \ref{tab:imageregions}. For better visualization, we normalize the results.
In Fig. \ref{fig:imageregions}, we can observe that with the increment of the image regions, the performance of all three tasks increases and then decreases. Moreover, selecting 5 image regions is optimal for VQA and VE. For VCR, 12 image regions are optimal. Visual fine-grained information can help CLIP and play an important role in the zero-shot prediction since it provides fine details of the image but more image regions after a certain point will result in a decrement. Too many image regions will introduce irrelevant visual information. In our experiments, we select 5 regions for VQA and SNLI-VE, and 12 regions for VCR.

\section{Conclusion}
\label{sec:conclusion}

In this work, we propose a unified and fine-grained approach for vision-language tasks including VQA, SNLI-VE, and VCR. We outperform previous CLIP-based methods for zero-shot VQA. Plus, we are the first to empirically study CLIP's zero-shot ability for SNLI-VE and VCR, which achieves strong zero-shot performance. In addition to the benchmark comparison, we conduct extensive ablation studies confirming the significance of visual and textual fine-grained information and the generalizability of our proposed method.

\section*{Limitations}


Although our proposed method is effective in three vision-language tasks, we still have some limitations. Firstly, we utilize T5 to convert the question-answering format into the declarative sentence in VQA and it works well in most cases, but it still faces out-of-coverage problems, which will affect the following zero-shot prediction of CLIP. We need to design more rules for these special cases for better conversion. Secondly, our clustering algorithm for SNLI-VE can achieve strong zero-shot performance, but the clustering centroids are close to each other and the algorithm is sensitive to these centroids. The robustness of this algorithm should be improved. What's more, we leverage Faster-RCNN in visual fine-grained information extraction, so the detectable object attributes and classes are constrained in a relatively limited object set of Faster-RCNN, which may hinder further improvement from visual fine-grained information. The Faster-RCNN can be replaced with a better vision module. Besides, since we only utilize CLIP in our paper, we can explore the zero-shot ability of other contrastive pre-training models in future work.




\section*{Ethics Statement}

There are many large-scale pre-trained models used in our paper like OFA, T5, RoBERTa, and CLIP. Our method relies heavily on CLIP, which is pre-trained on approximately 400M image-text pairs crawled from the Internet. Since the pre-training dataset is noisy, CLIP is likely to have potential racial and gender bias. Therefore, if someone finds our work interesting and would like to use it in a specific environment, we suggest the user check the potential bias before application. We think one advantage of our work is we only utilize existing pre-trained models and we don't need to train any new models. Compared to the energy-consuming model training, our method can be more environmentally friendly.


\section*{Acknowledgements}

We thank anonymous reviewers for their comments.
This work is supported by the DARPA MCS program under Cooperative Agreement N66001-19-2-4032. 

\bibliography{anthology,custom}
\bibliographystyle{acl_natbib}

\clearpage 

\appendix

\section{Appendix}
\label{sec:appendix}


\subsection{Data Statistics}\label{sec:statistics}

\begin{table}[h]\scriptsize
\setlength\tabcolsep{5pt} 
\centering
\begin{tabular}{lc|cc}
\toprule
\multicolumn{2}{c|}{VLU Tasks} & Number of Questions & Number of Images \\ 
\midrule
\multirow{4}{*}{VQAv2} & Yes/No & 80541 & - \\ 
 & Number & 28134 & - \\ 
 & Other & 105679 & - \\
 & All & 214354 & 40504 \\
 \midrule
 \multirow{4}{*}{SNLI-VE} & C & 5939 & - \\
 & N & 5960 & - \\
 & E & 5959 & - \\
 & All & 17858 & 1000 \\
 \midrule
 VCR & - & 26534 & 9929 \\
\bottomrule
\end{tabular}
\caption{\label{tab:statistics} 
Statistics of VQAv2, SNLI-VE, and VCR (C: Contradiction, N: Neutral, E: Entailment, VLU: Vision-Language Understanding)
}
\end{table}

Following previous work, we use the val2014 split of VQAv2. In zero-shot SNLI-VE and VCR, we use the validation set.



\subsection{Answer filtering for VQA}
\label{sec:answerfiltering}

\textbf{Answer filtering.} As in TAP-C \citep{song2022clip}, we first manually design the demonstrations and employ T5 to convert the question-answering format of VQA into the declarative template with the \textit{<extra\_id\_0>} token. Then, we input the concatenation of demonstrations and declarative statements converted from question-answering format with the \textit{<extra\_id\_0>} token into the T5 encoder. Next, encoded features from the T5 encoder and answer candidates are input into the T5 decoder. At the end of the T5 decoder, it will calculate the probability of each answer candidate. We select Top $K$ answers to replace \textit{<extra\_id\_0>} token in the template to generate $K$ prompts, which will be fed into the CLIP text encoder.

\textbf{Setting of hyperparameter $K$.} Since we employ answer filtering to select top $K$ answers, $K$ is a significant hyperparameter. In Tab. \ref{tab:topanswers}, we show how the zero-shot performance of VQA \textit{Number} and \textit{Other} type varies with the increment of selected top $K$ answers. We carry out six and seven experiments on these two types. We can observe that with the increment of $K$, the performance first increases and then decreases. When $K$ is small, many correct answers are directly removed by T5, which makes it impossible for CLIP to choose the right answer. Conversely, if $K$ is very big, there are too many answers, which are likely to disturb CLIP's zero-shot prediction. In our experiments, we select the top 10 answers in VQA \textit{Other} type and the top 4 answers in VQA \textit{Number} type.


\begin{algorithm}[H]
    \caption{Pseudocode of clustering algorithm} 
    \label{alg:ve} 
    \begin{algorithmic}[1] 
        \REQUIRE $\mathbb{V}$: CLIP image encoder, $\mathbb{T}$: CLIP text encoder, $I$: all images in SNLI-VE val split, $H$: all hypotheses in SNLI-VE val split, $N$: the number of samples in SNLI-VE val split;
        \ENSURE \textit{centroid}.
        \STATE {\textbf{dictionary} \textit{centroid} initialized to 0}
        \STATE {\textbf{array} \textit{scores} initialized to 0}
        \STATE {\textcolor{blue}{// use CLIP to calculate dot product}}
        \FOR {$ i = 0 $; $ i < N $; $ i ++ $ }
            \STATE {\textit{scores[i]} = $\mathbb{T}(\mbox{\textit{H[i]}}) \cdot \mathbb{V}(\mbox{\textit{I[i]}})$}
        \ENDFOR
        \STATE {\textcolor{blue}{// obtain three centroids}}
        \STATE {\textit{scores} = sort(\textit{scores}) \textcolor{blue}{// ascending order}}
        \STATE {\textcolor{blue}{// C: Contradiction, N: Neutral, E: Entailment}}
        \STATE {\textit{centroid[C]} = sum(\textit{scores[:N/3]}}) / (N/3)
        \STATE {\textit{centroid[N]} = sum(\textit{scores[N/3:2N/3]}}) / (N/3)
        \STATE {\textit{centroid[E]} = sum(\textit{scores[2N/3:]}}) / (N/3)
    \end{algorithmic} 
\end{algorithm}

\subsection{More ablation studies of VQAv2}

Since previous work uses CLIP RN50x16 to do zero-shot VQA, we also conduct ablation studies on it. Results can be found in Tab. \ref{tab:ablation2}

\subsection{Clustering algorithm and centroids of SNLI-VE}
\label{sec:clustering}


Algo. \ref{alg:ve} is utilized in zero-shot SNLI-VE. After running the Algo. \ref{alg:ve}, we can obtain three clustering centroids. In fact, we can cache the centroids in advance. In order to achieve better performance, we tune the centroids, which are reported in Tab. \ref{tab:centroid}. The effectiveness of Algo. \ref{alg:ve} is based on the relatively even data distribution. K-Means \footnote{\href{https://en.wikipedia.org/wiki/K-means_clustering}{https://en.wikipedia.org/wiki/K-means\_clustering}} can also be utilized here but it also requires a relatively even data distribution. In the validation split of SNLI-VE, we have 17858 samples, which are not divisible by 3. However, we can assume there are 5952, 5953, and 5953 samples in entailment, neutral, and contradiction category respectively.

\subsection{How \# image regions affect performance}
\label{sec:imageregions}

The full results are reported in Tab. \ref{tab:imageregions}. They are values before normalization in Fig. \ref{fig:imageregions}. Through the table and figure, we can see how selected $N$ images affects the zero-shot performance.


\subsection{Zero-shot learning by only using textual fine-grained information}

We think it is interesting to investigate the zero-shot performance if we only use textual fine-grained information. We only exploit the language model to accomplish zero-shot prediction in all three vision-language tasks. All results are shown in Tab. \ref{tab:lanfine}. In VQA, we use T5-large (for answer filtering) and RoBERTa-large. In SNLI-VE and VCR, we only utilize RoBERTa-large. Visual information is not considered and textual fine-grained information includes the image caption and question in this experimental setting. All results show that only using textual fine-grained information can achieve fair performance. (Note: We can notice that only using ground truth textual fine-grained information in SNLI-VE can surpass baseline performance. It is because the relation between the ground truth caption and hypothesis is well annotated in SNLI \citep{bowman2015large})

\begin{table*}
\centering
\begin{tabular}{!{\color{black}\vrule}c!{\color{black}\vrule}c!{\color{black}\vrule}c!{\color{black}\vrule}c!{\color{black}\vrule}c!{\color{black}\vrule}c!{\color{black}\vrule}c!{\color{black}\vrule}c!{\color{black}\vrule}c!{\color{black}\vrule}c!{\color{black}\vrule}c!{\color{black}\vrule}} 
\hline
\begin{tabular}[c]{@{}c@{}}Answer \\Type\end{tabular} & 2 & 3 & 4 & 5 & 9 & 10 & 11 & 20 & 40 & 200 \\ 
\hline
Number & 25.14 & 25.27 & \textbf{25.36} & 24.41 & - & 21.60 & - & 18.36 & - & - \\ 
\hline
Other & - & - & - & 14.98 & 16.60 & \textbf{16.65} & 16.61 & 15.54 & 14.93 & 10.87 \\
\hline
\end{tabular}
\caption{\label{tab:topanswers} 
How \# selected answers affect VQA
}
\end{table*}

\begin{table*}\scriptsize
\setlength\tabcolsep{1pt} 
\renewcommand{\arraystretch}{1.1} 
\centering
\begin{tabular}{l|cc|cccccccc} 
    \toprule
    Methods & \multicolumn{2}{c|}{VLU Tasks} & Baseline & w/ Question & \makecell[c]{w/ Image Region \\ (Faster-RCNN)} & \makecell[c]{w/ Image Caption \\ (OFA)} & \makecell[c]{w/ All \\ (Generation)} & \makecell[c]{w/ Image Caption \\ (GT)} & \makecell[c]{w/ Image Region \\ (GT)} & \makecell[c]{w/ All \\ (GT)} \\
    \midrule
    \multirow{4}{*}{\makecell[l]{UniFine-Base \\ w/ CLIP$_{\mbox{\tiny Res50x16}}$}} & \multirow{4}{*}{VQA} & Yes/No & 68.85 & 68.90 \textcolor{cyan}{[+0.05]} & 68.90 \textcolor{cyan}{[+0.05]} & 69.63 \textcolor{cyan}{[+0.78]} & 69.69 \textcolor{cyan}{[+0.84]} & 69.59 \textcolor{cyan}{[+0.74]} & - & 69.66 \textcolor{cyan}{[+0.81]} \\
    & & Number & 25.85 & 26.20 \textcolor{cyan}{[+0.35]} & 26.34 \textcolor{cyan}{[+0.49]} & 29.26 \textcolor{cyan}{[+3.41]} & 29.61 \textcolor{cyan}{[+3.76]} & 28.91 \textcolor{cyan}{[+3.06]} & - & 29.33 \textcolor{cyan}{[+3.48]} \\
    & & Other & 16.74 & 17.36 \textcolor{cyan}{[+0.62]} & 18.60 \textcolor{cyan}{[+1.86]} & 17.86 \textcolor{cyan}{[+1.12]} & 19.85 \textcolor{cyan}{[+3.11]} & 17.64 \textcolor{cyan}{[+0.90]} & - & 19.85 \textcolor{cyan}{[+3.11]} \\
    & & All & 37.53 & 37.90 \textcolor{cyan}{[+0.37]} & 38.53 \textcolor{cyan}{[+1.00]} & 38.82 \textcolor{cyan}{[+1.29]} & 39.87 \textcolor{cyan}{[+2.34]} & 38.65 \textcolor{cyan}{[+1.12]} & - & 39.82 \textcolor{cyan}{[+2.29]} \\
    \bottomrule
\end{tabular}
\caption{\label{tab:ablation2} 
Ablation Studies of zero-shot VQA. (GT: Ground Truth)
}
\end{table*}

\begin{table*}[!htbp]\small
\setlength\tabcolsep{4pt} 
\renewcommand{\arraystretch}{1.2} 
\centering
\begin{tabular}{!{\color{black}\vrule}c!{\color{black}\vrule}c!{\color{black}\vrule}c!{\color{black}\vrule}c!{\color{black}\vrule}c!{\color{black}\vrule}c!{\color{black}\vrule}c!{\color{black}\vrule}} 
\hline
\multirow{3}{*}{Answer Types} & \multicolumn{6}{c!{\color{black}\vrule}}{Methods} \\ 
\cline{2-7}
& \multicolumn{2}{c!{\color{black}\vrule}}{CLIP ViT-B/16} & \multicolumn{2}{c!{\color{black}\vrule}}{CLIP ViT-L/14@336px} & \multicolumn{2}{c!{\color{black}\vrule}}{RoBERTa-large}                                                                                        \\ 
\cline{2-7}
& Baseline & w/ Image Region  & Baseline & w/ Image Region & \begin{tabular}[c]{@{}c@{}}Only Image Caption \\(Generation)\end{tabular} & \begin{tabular}[c]{@{}c@{}}Only Image Caption \\(GT)\end{tabular} \\ 
\hline
Contradiction & 0.23 & 0.47 & 0.17 & 0.37 & 0.22 & 0.29 \\ 
\hline
Neutral & 0.26 & 0.54 & 0.22 & 0.45 & 0.34 & 0.48 \\ 
\hline
Entailment & 0.27 & 0.55 & 0.23 & 0.46 & 0.43 & 0.60 \\
\hline
\end{tabular}
\caption{\label{tab:centroid} 
Clustering centroids of SNLI-VE (GT: Ground Truth).
}
\end{table*}

\begin{table*}[!htbp]\scriptsize
\setlength\tabcolsep{5pt} 
\centering
\begin{tabular}{l|cccccccccccccccc}
\toprule
Task & 0 & 1 & 2 & 3  & 4  & 5  & 6 & 7 & 8 & 9 & 10 & 11 & 12 & 13 & 14 & 15 \\ 
\midrule
\makecell[l]{VQA \\ Yes/No} & 68.61 & 68.67 & 68.72 & 68.75 & 68.76 & \textbf{68.80} & 68.77 & 68.74 & 68.73 & 68.74 & 68.73 & 68.71 & 68.72 & 68.70 & 68.67 & 68.66  \\ 
\midrule
VE & 47.37 & 47.42 & 47.39 & 47.40 & 47.45 & \textbf{47.54} & 47.49 & 47.45 & 47.36 & 47.40 & 47.11 & 47.08 & 47.13 & 47.14 & 47.14 & 47.12 \\ 
\midrule
\makecell[l]{VCR \\ Q2A} & 53.24 & 52.68 & 53.06 & 53.27 & 53.42 & 53.46 & 53.61 & 53.69 & 53.80 & 53.81 & 53.82 & 53.90 & \textbf{53.93} & 53.91 & 53.83 & 53.81 \\
\bottomrule
\end{tabular}
\caption{\label{tab:imageregions} 
How \# image regions affect the performance of different tasks
}
\end{table*}

\begin{table*}\small
\centering
\begin{tabular}{l|cccc|cccc|cc} 
\toprule
\multirow{2}{*}{Methods} & \multicolumn{4}{c|}{VQA Answer Types} & \multicolumn{4}{c|}{SNLI-VE Answer Types} & \multicolumn{2}{c}{VCR Tasks} \\ 
\cmidrule{2-11} & Yes/No & Number & Other & All & C & N & E & All & Q2A & QA2R \\ 
\midrule
Baseline w/ CLIP$_{\mbox{\tiny ViT-B/16}}$ & 68.61 & 25.36 & 16.65 & 37.33 & 67.59 & 18.66 & 55.92 & 47.37 & 53.24 & 46.51 \\ 
\midrule
\makecell[l]{Only textual fine-grained \\information (Generation)} & 67.13  & 28.77  & 13.44 & 35.64 & 64.83 & 17.32 & 56.19 & 46.09 & 35.13 & 41.93 \\ 
\midrule
\makecell[l]{Only textual fine-grained \\information (GT)} & 67.58  & 28.22  & 13.36 & 35.69 & 68.72 & 24.03 & 63.00 & 51.89 & 32.69 & 40.21 \\
\bottomrule
\end{tabular}
\caption{\label{tab:lanfine} 
Zero-shot performance when only using textual fine-grained information. (C: Contradiction, N: Neutral, E: Entailment, GT: Ground Truth)
}
\end{table*}

\end{document}